\documentclass[runningheads]{llncs}
\usepackage{graphicx}
\usepackage{siunitx}

\usepackage{tikz}
\usepackage{comment}
\usepackage{amsmath,amssymb} 
\usepackage{color}
\usepackage{multirow}

\begin{document}
\pagestyle{headings}
\mainmatter
\def\ECCVSubNumber{66}  

\title{Deformable Kernel Convolutional Network for Video Extreme Super-Resolution} 

\titlerunning{Deformable Kernel Convolutional Network for Video Extreme SR}

\author{Xuan Xu\inst{1}\orcidID{0000-0003-1057-3286}\and
Xin Xiong\inst{2} \and
Jinge Wang\inst{1} \and \\ 
Xin Li\inst{1}\inst{,\dagger}\orcidID{0000-0003-2067-2763}
}
\authorrunning{X. Xu et al.}
%
\institute{West Virginia University, Morgantown WV 26505, USA \and
Huazhong University of Science and Technology, Wuhan 430074, China
\email{\{xuxu,jnwang1\}@mix.wvu.edu;}
\email{xiong\_xin@hust.edu.cn;}\\
\email{xin.li@mail.wvu.edu}
}

\maketitle

\begin{abstract}

Video super-resolution, which attempts to reconstruct high-resolution video frames from their corresponding low-resolution versions, has received increasingly more attention in recent years. Most existing approaches opt to use deformable convolution to temporally align neighboring frames and apply traditional spatial attention mechanism (convolution based) to enhance reconstructed features. However, such spatial-only strategies cannot fully utilize temporal dependency among video frames. In this paper, we propose a novel deep learning based VSR algorithm, named Deformable Kernel Spatial Attention Network (DKSAN).  Thanks to newly designed Deformable Kernel Convolution Alignment (DKC\_Align) and Deformable Kernel Spatial Attention (DKSA) modules, DKSAN can better exploit both spatial and temporal redundancies to facilitate the information propagation across different layers. We have tested DKSAN on AIM2020 Video Extreme Super-Resolution Challenge to super-resolve videos with a scale factor as large as 16. Experimental results demonstrate that our proposed DKSAN can achieve both better subjective and objective performance compared with the existing state-of-the-art EDVR on Vid3oC and IntVID datasets. 
    
\end{abstract}

\section{Introduction}

\let\thefootnote\relax\footnotetext{\inst{\dagger}Corresponding author}

Video Super-Resolution (VSR) refers to the task of reconstructing high-resolution (HR) video frames from their corresponding low-resolution (LR) observation data. Similar to image super-resolution, VSR aims at faithful recovery of important image structures (e.g., edges and textures) and has been widely used in practical applications from video surveillance \cite{seibel2017eyes} and high-definition Television (HDTV) \cite{matsuo2017super} to video coding and  streaming \cite{umeda2018hdr}. Existing VSR research can be mainly classified into two subfields, enhancing spatial super-resolution and enhancing temporal super-resolution. The former focuses on super-resolving LR video frames to approximate HR video frames to improve visual quality of video; while the later refers to interpolate new frames between neighboring frames for the purpose of increasing video frame rate (e.g., from 30fps to 60fps). Different from Single Image Super-Resolution (SISR) which only needs to consider the information from spatial domain, both spatial and temporal dependencies have to be utilized by VSR algorithms in order to optimize their performance. In particular, how to effectively exploit temporal redundancy by motion compensation techniques has remained one of the key technical challenges in the task of VSR. 

In order to explore the potential benefit from temporal information of VSR, several existing approaches \cite{farsiu2004fast},\cite{liu2013bayesian},\cite{ma2015handling},\cite{wang2018video} have used a sequence of consecutive LR frames (including one reference frame and several neighboring frames) as inputs to reconstruct the HR frame corresponding to the reference LR frame. To better exploit temporal dependency among multiple LR frames, the consecutive frames need to be aligned before the reconstruction of the HR frame. One of the most popular motion estimation methods, optical-flow estimation \cite{optical_flow}, is often considered and has been adopted by several VSR approaches \cite{liu2017robust},\cite{sajjadi2018frame},\cite{ESPCN_video}. However, VSR based on rigid motion estimation has to suffer from the potential problem arising from misalignment. For example, it is well known that there are two plagues with optical flow estimation: occlusion and aperture problems \cite{tekalp2015digital}. VSR based on incorrect motion estimation results may introduce undesired blurring and misregistration artifacts to the reconstructed HR frames.

In view of the weakness of rigid motion estimation approaches, alternative methods - namely deformable motion estimation - have been proposed as well.
Recently, {\it deformable convolution} \cite{deformableV1},\cite{deformableV2} has become more and more popular as a supplementary module to video frame alignment. Several VSR works such as \cite{EDVR},\cite{TDAN},\cite{deformable_non_local} have already successfully applied varying forms of deformable convolution alignment module to temporally align neighboring frames with respect to the reference frame, which demonstrates improved motion compensation when compared with optical-flow-based methods. However, existing deformable alignment modules still learn the motion parameters via several standard convolution layers with fixed kernel configurations, which can not extract accurate motion information especially in the presence of large and deformable motion (e.g., in sport video). By contrast, deformable kernels \cite{deformablekernel} can adapt effective receptive fields \cite{luo2016understanding} (i.e., the support of filters) by weighting the per-pixel contribution, which is expected to be capable of characterizing more sophisticated motion information.

In this paper, we propose a novel Multi-Frame based Deformable Kernel Spatial Attention Network (DKSAN) for video {\it extreme} super-resolution (the upscaling factor is as large as 16). Inspired by EDVR \cite{EDVR} which applies deformable convolution \cite{deformableV2} to temporally align neighboring frames with reference frame, we have designed a new module not based on optical flow estimation, called Deformable Kernel Convolution Alignment (DKC\_Align) module, to enhance deformable convolution \cite{deformableV2}. The key idea is to combine deformable kernel with deformable convolution to extract and improve not only global but also local edge and texture features while aligning the neighboring frames with respect to the reference frame. Moreover, we have developed a Deformable Kernel Spatial Attention (DKSA) module to further enhance the spatial details of reconstructed feature maps, which extends the previous spatial attention works such as \cite{EDVR},\cite{Group},\cite{SCAN}. The novelty of DKSA module lies in that the deformable kernel \cite{deformablekernel} can better represent spatially-localized edge and texture features which are often important for the task of VSR than conventional convolution based spatial attention.

\section{Related Works}
Unlike image super-resolution which deals with reconstructing missing information in the spatial domain only, VSR has to not only reconstruct the missing high-frequency information in the spatial domain but also consider the motion-related consistency across different video frames in the temporal domain. In this section, we briefly review existing VSR approaches based on multi-frame such as \cite{EDVR},\cite{TDAN},\cite{Zooming},\cite{Group},\cite{deformable_non_local},\cite{ESPCN_video}, optical flow \cite{optical_flow} alignment and deformable convolution \cite{deformableV2} alignment.

\begin{figure}
    \centering
    \includegraphics[width=12cm]{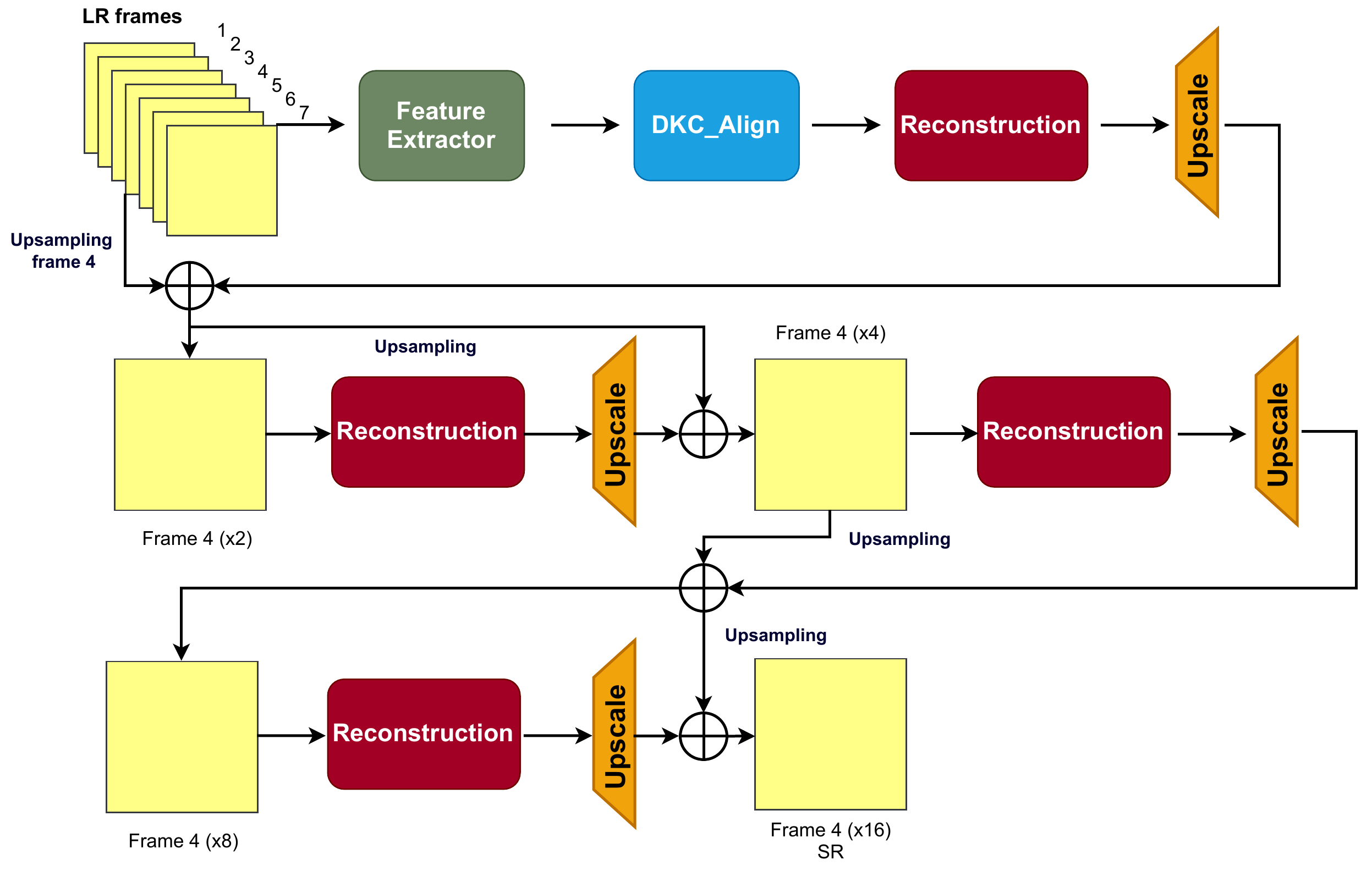}
    \caption{Overview of DKSAN, $\oplus$ denotes element-wise sum.}
    \label{fig:overview_video}
\end{figure}

\subsection{Video Super-Resolution}
One of the early works of applying optical-flow to VSR problems in order to utilize temporal and spatial information is \cite{liao2015video}. In this work, a draft-ensemble strategy was introduced to use two robust optical flow methods: TV-$l_1$ flow and MDP flow to overcome the difficulty with large motion variation and then combine SR drafts via a deep convolutional neural network to generate the final SR result. Later, \cite{kappeler2016video} proposed to use optical-flow to estimate motion compensation of consecutive LR frames and wrapped them as inputs of the {CNN} to generate SR frames. Those two-stage approaches are not optimal solution since they separate the motion compensation from frame reconstruction. To explore potential benefits of end-to-end learning architecture for VSR problem, a novel end-to-end deep {CNN} to joint train the estimation of optical flow and spatio-temporal networks called ESPCN was developed in  \cite{ESPCN_video}. In \cite{tao2017detail}, a new layer called sub-pixel motion compensation (SPMC) was introduced to handle inter-frame motion alignment; it also applied a ConvLSTM \cite{xingjian2015convolutional} architecture for reconstruction and testing. Another work \cite{RBPN} proposed a recurrent back-projection network (RBPN) with encoder-decoder mechanism to extract spatial and temporal information. In \cite{jo2018deep}, dynamic upsampling filters (DUF) was developed to avoid use the explicit motion compensation by computing pixels of local spatio-temporal neighbors of LR frames to learn implicit motion compensation.  Most recently,  a novel temporal group attention (TGA) framework \cite{Group} was proposed to group the input frames (7 frames) as three groups then generate temporal spatial attention maps to reconstruct the missing details in the reference frame. Another recent work \cite{xue2019video} proposed to learn self-supervised motion representation, task-oriented flow (TOFlow), instead of fix optical flow as the motion compensation module for VSR problem. 

\subsection{Deformable Convolution}

The inherent limitation with traditional CNNs is the capability of modeling geometric transformations because of their fixed kernel shape. Although dilated convolution can alleviate this limitation to some degree, it is still difficult for standard fixed-shape convolutional kernels to align the key points or salient features in the input images. To solve this problem, a deformable convolution network has been developed in \cite{deformableV1},\cite{deformableV2} to improve the capability of modeling geometric transformations by adding flexible and learnable offsets. By acquiring information from other field rather than fixed local area,
deformable convolution networks have been widely used by high-level vision tasks such as object detection \cite{bertasius2018object} and segmentation \cite{deformableV1}. Inspired by \cite{deformableV2}, a recent work \cite{TDAN}  proposed a temporally-deformable alignment network (TDAN) to adapt deformable convolution to align the consecutive LR input frames at the feature level. Along this line of research, EDVR \cite{EDVR} designed a more aggressive alignment approach, PCD align module, to align the neighboring LR frames at different scale levels; also they proposed a temporal and spatial attention fusion module to future enhance important features. Another recent work \cite{Zooming} proposed a novel space-time video super-resolution framework to utilize deformable convolution and deformable ConvLSTM module to achieve temporal and spatial super-resolution at the same time. Most recently, \cite{deformable_non_local} introduced another deformable convolution based {VSR} framework called deformable non-local network (DNLN) with non-local attention module and hierarchical feature fusion block to enhance the global details between neighboring frames and references. Those deformable alignment based methods have shown better performance than optical-flow based networks.

\begin{figure}
    \centering
    \includegraphics[width=12cm]{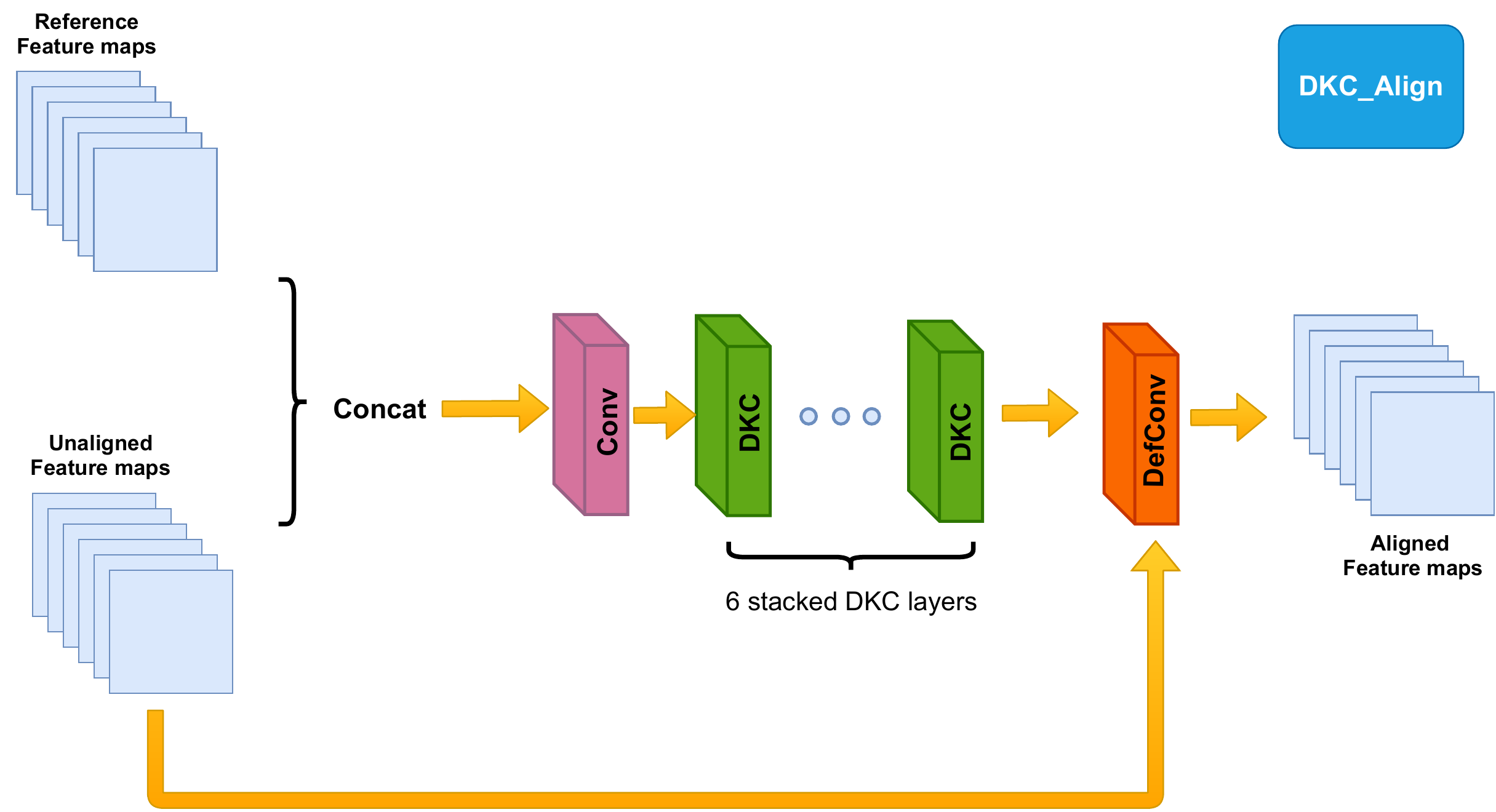}
    \caption{Overview of DKC\_Align module. Conv means convolution layer, DKC means deformable kernel convolution layer and DefConv stands for deformable convolution layer.}
    \label{fig:DKC_Align}
\end{figure}

\section{Proposed Methodology}
\begin{figure}[ht]
    \centering
    \includegraphics[width=12.cm]{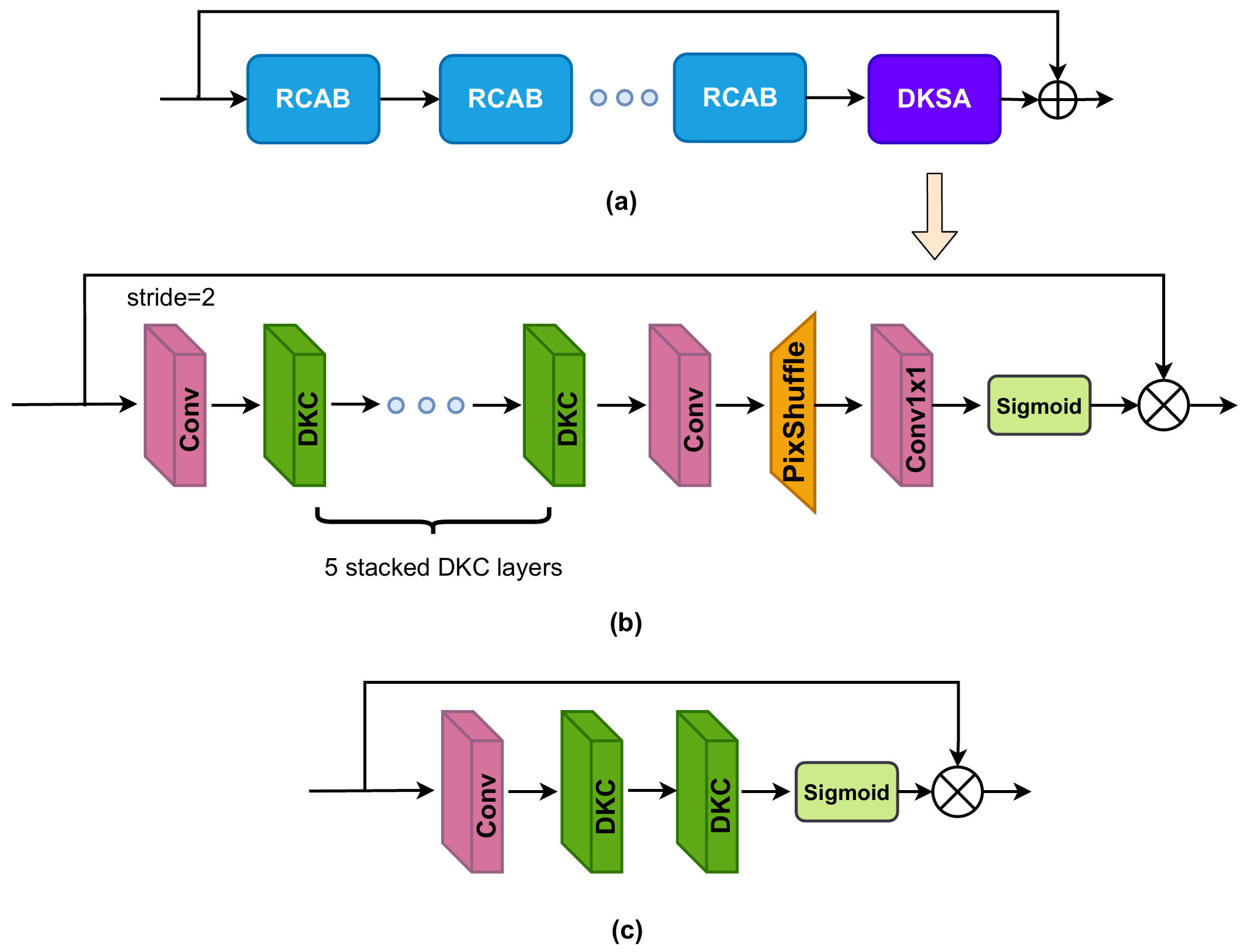}
    \caption{Overview of reconstruction module;  DKSA is deformable kernel spatial attention module shown in (b); a light version of DKSA is shown in (c); $\oplus$ and $\otimes$ denote element-wise sum and element-wise product, respectively.}
    \label{fig:recon_video}
\end{figure}

The design of DKSAN network can be presented in the order of top-down hierarchy: DKSAN network (Fig.~\ref{fig:overview_video}) $\rightarrow$ DKC\_Align subnetwork (Fig.~\ref{fig:DKC_Align})$\rightarrow$ reconstruction module (Fig.~\ref{fig:recon_video}).

\subsection{Overview: Deformable Kernel Spatial Attention Network}
For multi-frame based VSR, we are given a group of $2N+1$ consecutive LR frames $I^{LR}_{T} = \{I^{LR}_{r-N}, \dots, I^{LR}_{r-1}, I^{LR}_{r}, I^{LR}_{r+1}, \dots, I^{LR}_{r+N} \}$, where $I^{LR}_{r}$ is denoted as frame at the center or reference frame and $I^{LR}_{r-N}$ or $I^{LR}_{r+N}$ are the neighboring frames of $I^{LR}_{r}$. The goal of multi-frame based VSR is to reconstruct a HR frame $\hat{Y}_{r}$ from the LR sequence of $I^{LR}_{T}$ by exploiting both spatial and temporal redundancies in the sequence. The overall diagram of our proposed networks {DKSAN} is shown in Fig.~\ref{fig:overview_video}. It mainly includes four parts: {\it feature extraction, DKC\_Align module, reconstruct module}, and {\it upscale module}. Different from traditional deep learning based multi-frame VSR architectures, this work aims at super-resolving the LR videos at the extreme cases (e.g., with the scaling factor of 16). Due to large scaling factor constraint, it is difficult to upscale the LR feature maps to the target HR ones directly. One-time upscaling approaches such as \cite{EDVR},\cite{TDAN},\cite{Group} tend to introduce undesired blurring and artifacts to super-resolved HR video frames. 

To address this issue, we propose to construct a cascade of upscaling building blocks to iteratively super-resolve LR features several times (four times to reach the factor of $16=2^4$). Thanks to the cascade  architecture, the LR frames can be super-resolved progressively to reconstruct the unknown HR frames more accurately than previous one-shot approaches.
The whole problem of VSR can be formulated as follows:
\begin{equation}
    \hat{Y}_{r} = \mathbb{F}(I^{LR}_{T})
    \label{eq:ch4_1}
\end{equation}
where $I^{LR}_{T}$ denotes the consecutive LR frames and $\hat{Y}_{r}$ denotes the super-resolved reference frame $I^{LR}_{r}$. In particular, we extract the preliminary features of all input frames through the feature extraction which is stacked by several resblocks \cite{EDVR} without batch normalization layers. This procedure can be represented by:
\begin{equation}
    F_{fea} = E_{res}(I^{LR}_{T})  
\end{equation}
where $E_{res}$ denotes the preliminary feature extraction, the output $F_{fea}$ is the extracted feature maps for all input frames. Let define $F_{n}$ is the neighboring feature, and $F_{r}$ is the reference feature separated from $F_{fea}$. To align the neighboring feature and the reference feature with the proposed DKC\_Align module $E_{DKC\_Align}$, we have
\begin{alignat}{2}
   & F_{Align} = E_{DKC\_Align}(F_{n}, F_{r}) \\
   & F_{fusion} = \textbf{W}_E(F_{Align}) 
\end{alignat}
where $n \in [t-N, t+N]$ and $n \not= r$, $F_{Align}$ is the concatenated aligned feature maps for each neighboring frame feature with reference frame feature. The details about this alignment module will be elaborated in section \ref{alignment}; $\textbf{W}_E \in \mathbb{R}^{1 \times 1 \times C}$ is a $1\times1$ Conv layer. Conceptually similar to encoder-decoder configuration \cite{cheng2019encoder},\cite{RBPN}, the aligned feature $F_{Align}$ (encoder outputs) will be fed to the reconstruction module and upscale module for the first-level upscaling (decoder) operation:
\begin{equation}
    \hat{Y}^{level1}_{r} = U_{1}(E_{Recon1}(F_{fusion})) + B_{2\times}(I^{LR}_r)
    \label{eq:recon1}
\end{equation}
where $E_{Recon1}$ denotes the first level reconstruction module, $U_{1}$ is the first level upscaling module and $B_{2\times}$ stands for the Bicubic interpolation with scale factor of 2; $\hat{Y}^{level1}_{r}$ is the $2\times$ SR frame. Finally, to get the extreme super-resolved frame $\hat{Y}_{r}$, we repeat another 3 times of reconstruction operation:
\begin{alignat}{3}
    & \hat{Y}^{level2}_{r} = U_{2}(E_{Recon2}(E_{2}(\hat{Y}^{level1}_{r}))) + B_{2\times}(\hat{Y}^{level1}_{r}) \\
    & \hat{Y}^{level3}_{r} = U_{3}(E_{Recon3}(E_{3}(\hat{Y}^{level2}_{r}))) + B_{2\times}(\hat{Y}^{level2}_{r}) \\
    & \hat{Y}_{r} = U_{4}(E_{Recon4}(E_{4}(\hat{Y}^{level3}_{r}))) + B_{2\times}(\hat{Y}^{level3}_{r}) 
\end{alignat} 
where $E_{2}, E_{3}, E_{4}$ are the preliminary feature extractors for each level; $U_{2}, U_{3}, U_{4}$ denote the upscaling module for each corresponding level, respectively. The details about the reconstruction module are described in section \ref{reconstruct} including the DKSA module.

\subsection{Deformable Kernel Alignment Module} \label{alignment}
Different from previous VSR works which applied optical flow to align neighboring frames with reference frame, \cite{TDAN} and \cite{EDVR} introduced to utilize modulated deformable convolution \cite{deformableV2} to temporally align the given consecutive frames in order to add temporal information to VSR frameworks. 

\subsubsection{Deformable Convolution and Deformable Kernel} \label{sec:dkc}
Inspired by \cite{deformablekernel},\cite{EDVR}, we propose a new alignment module, DKC\_Align, to combine the deformable kernel \cite{deformablekernel} and deformable convolution \cite{deformableV2} as shown in Fig.~\ref{fig:DKC_Align}. First, let $F_{n}$ and $F^{align}_{n}$ denote the input and output feature maps (not the reference frame feature), $\textbf{W}_k$ represents the weight kernel and $p_k$ is the pre-specified offsets for the $k$-th location ($K$ is the total sampling location), then the modulated deformable convolution can be described as follows:
\begin{equation}
    F^{align}_{n}(p) = \sum_{k\in K}\textbf{W}_{k}\cdot F_{n}(p+p_k+\Delta p_k)\cdot \Delta m_k
\end{equation} 
where $F^{align}_{n}(p)$ and $F_{n}(p)$ indicate the feature location $p$ from $F_{n}^{align}$ and $F_{n}$, $\Delta p_k$ and $\Delta m_k$ stand for the learnable offset and the modulation scalar, respectively. With $\Delta p_k$ and $\Delta m_k$, the convolution will get the ability to be irregularly dilated to work with important feature points without the shape limitation of conventional convolution. Such process of deformable convolution can be regarded as a strategy of adapting the local receptive field to a support of arbitrary shape.

To get $\Delta p_k$ and $\Delta m_k$ and align the neighboring feature with reference feature in particular, we first concatenate the neighboring frame feature and the reference frame feature then fuse them with one Conv2D layer and fed them into several deformable kernel layers:
\begin{equation}
    \Delta P_{n}, \Delta M_{n} = \mathbb{D}(f([F_n, F_r])) , n \in [t-N, t+N], n \not = r
\end{equation}
where $f$ denotes the one Conv2d layer to fuse $F_n$ and $F_r$, $\mathbb{D}$ represents the deformable kernel convolution layer. To formally express deformable kernel convolution layer $\mathbb{D}$, let $ \Delta k $ denote a learnable offset of the kernel $\textbf{W}$, then deformable kernel convolution layer can be formulated as:
\begin{equation}
    \mathbb{D} = \sum_{k\in K}\textbf{W}_{k+\Delta k}\cdot f([F_n, F_r])(p+p_k+\Delta p_k), n \in [t-N, t+N], n \not = r
\end{equation}

According to \cite{deformablekernel}, deformable convolution can only adapt theoretical receptive fields by deforming the conventional convolution, but it cannot evaluate the contribution of each grid point. As a complementary tool to deformable convolution \cite{deformableV1},\cite{deformableV2}, deformable kernel \cite{deformablekernel} can weigh the contribution of each grid point to inform the network which point is more important (i.e., adaptive control of effective receptive fields). The advantage of combining deformable convolution with deformable kernel is to not only deform the convolution for extracting key grid points but also adaptively weigh the importance of each point (similar to the introduction of attention mechanism \cite{vaswani2017attention}). This way, the deformable convolution kernel layers will be more sensitive to the key feature points than traditional convolution layers and capable of extracting richer information to improve the alignment accuracy and reconstruction quality for our VSR task. Note that previous work such as EDVR \cite{EDVR} only studies the benefit of deformable convolution in VSR; while deformable kernel \cite{deformablekernel} was originally designed for high-level vision tasks such as object detection and classification. To the best of our knowledge, this work is the first to leverage of idea of combining deformable convolution with deformable kernel into the application of VSR.

\subsection{Reconstruction Module}  \label{reconstruct}
\begin{figure}
    \centering
    \includegraphics[width=5cm]{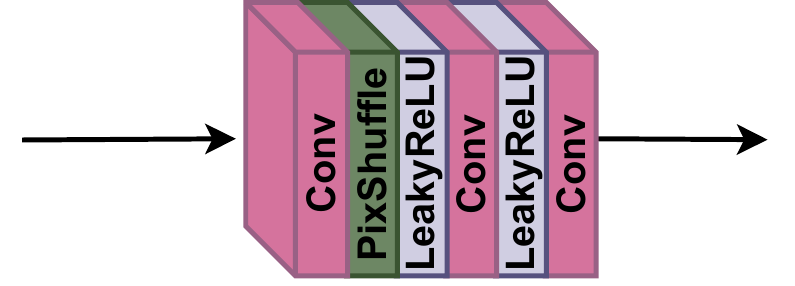}
    \caption{The details of upscale module, the last Conv layer has only 3 feature maps output in order to generate RGB color frame.}
    \label{fig:upscale}
\end{figure}

To get the super-resolved frame $\hat{Y}_{r}$, the output $F_{fusion}$ from the DKC\_Align module is fed into the reconstruction module. The reconstruction module includes several stacked RCAB blocks and the {DKSA} module (please refer to  Fig.~\ref{fig:recon_video} (a)):
\begin{equation}
    F_{recon} = E_{DKSA}(E_{RCABs}(F_{fusion})) + F_{fusion}
\end{equation}
where $F_{recon}$ is the final reconstruction features to be fed into upscale module (the architecture of upscale module is shown in Fig.~\ref{fig:upscale} which includes several Conv layers, PixelShuffle and LeakyReLU), $ E_{RCABs}$ and $E_{DKSA}$ denote the RCAB blocks and DKSA module. Note that RCAB module has the same structure as it proposed in RCAN \cite{RCAN} which includes resblock \cite{resnet} and channel attention mechanism \cite{hu2018squeeze},\cite{RCAN},\cite{JDSR}.   

\subsubsection{Deformable Kernel Spatial Attention Module}
In order to further calibrate output feature maps, we propose to construct a new Deformable Kernel based Spatial Attention (DKSA) module instead of traditional spatial attention mechanism. As shown in Fig.~\ref{fig:recon_video} (b), in DKSA, we first use one Conv layer to extract the output of the stacked RCAB blocks, then a couple of stacked Deformable Kernel Convolution (DKC) layers are placed to further extract key features from the naive feature map. As discussed in section \ref{sec:dkc}, deformable kernels can better measure the effective receptive field than standard convolution kernels. Therefore, DKSA can generate improved spatial attention maps to enforce networks pay more attention to important features such as edges and textures. Note that Fig.~\ref{fig:recon_video} (c) shows a light version of DKSA which is used by the level-1 reconstruction module.

\begin{table}
   
    \begin{center}
    \begin{tabular}{c|c|c|c|c}
    \hline
    \multirow{2}{*}{Video Name} &   \multirow{2}{*}{Scale}  &  Bicubic  & EDVR & DKSAN (ours)  \\
    \cline{3-5} & & PSNR (dB) & PSNR (dB) & PSNR (dB)\\
    \hline\hline
      050  & x16  & 25.36   & 26.75   & \textbf{29.17}   \\
      051  & x16  & 23.20  & 23.76   & \textbf{24.72}  \\
      052  & x16  & 20.57  & 20.92   & \textbf{21.61}  \\
      053  & x16  & 21.61 & 22.15   &  \textbf{22.63} \\
      054  & x16  & 20.08   & 20.56   & \textbf{21.15}  \\
      055  & x16  & 20.01   & 20.36  & \textbf{21.48}  \\
      056  & x16  & 21.44  & 21.33   & \textbf{22.54}  \\
      057  & x16  & 20.22   & 20.33   &  \textbf{21.66} \\
      058  & x16  & 19.55  & 19.80   &  \textbf{21.45} \\
      059  & x16  & 20.22   & 20.92   &  \textbf{21.90} \\
      060  & x16  & 20.13   & 20.30   & \textbf{21.38}  \\
      061  & x16  & 21.08   & 21.58   & \textbf{22.22}  \\
      062  & x16  & 21.54   & 21.58   & \textbf{23.12}  \\
      063  & x16  & 21.54   & 22.00  &  \textbf{23.26} \\
      064  & x16  & 20.46   & 21.04   &  \textbf{21.94} \\
      065  & x16  & 22.53   & 23.41   &  \textbf{24.80} \\
      \hline\hline
    Average & x16  & 21.22  & 21.67   &  \textbf{22.81} \\
    \hline
    Parameters &  -  &  -  & 20.6M & 29.5M \\
    \hline
    Runtime(s/f) & - &  - &  0.87  & 0.95  \\
 
    \hline
    \end{tabular}
    \end{center}
    \caption{Quantitative comparison on Vid3oC dataset for scaling factor of 16; s/f means seconds per frame. \textbf{Bold} font indicates the best result.}
    \label{tab:PSNR_vid3oc}
\end{table}

\begin{table}
   
    \begin{center}
    \begin{tabular}{c|c|c|c|c}
    \hline
    \multirow{2}{*}{Video Name} &   \multirow{2}{*}{Scale}  &  Bicubic  & EDVR & DKSAN (ours)  \\
    \cline{3-5} & & PSNR (dB) & PSNR (dB) & PSNR (dB)\\
    \hline\hline
      050  & x16  & 21.56   & 21.81   & \textbf{23.06}   \\
      051  & x16  & 23.02   & 24.13   & \textbf{24.92}  \\
      052  & x16  & 29.56   & 29.33   & \textbf{31.87}  \\
      053  & x16  & 24.05   & 24.51   &  \textbf{25.09} \\
      054  & x16  & 31.34   & 33.15   & \textbf{36.18}  \\
      055  & x16  & 24.39   & 25.01   & \textbf{26.88}  \\
      056  & x16  & 31.16   & 31.93   & \textbf{34.22}  \\
      057  & x16  & 34.35   & 35.20   &  \textbf{39.75} \\
      058  & x16  & 36.00   & 37.36   &  \textbf{38.15} \\
      059  & x16  & 30.49   & 31.37   &  \textbf{34.17} \\
      \hline\hline
    Average & x16  & 28.59   & 29.38   &  \textbf{31.43} \\
    \hline
    \end{tabular}
    \end{center}
    \caption{Quantitative comparison on IntVID dataset for scaling factor of 16. \textbf{Bold} font indicates the best result.}
    \label{tab:PSNR_ch4}
\end{table}

\section{Experimental Results}
In this section, we demonstrate the training and test datasets, network setting, training details, experimental results and ablation study of proposed video extreme super-resolution approach.

\subsection{Datasets} \label{datasets}
In this work, the training data we have used is Vid3oC \cite{vid3oc} dataset provided by AIM2020 Video Extreme Super-Resolution Challenge. The Vid3oC dataset includes 50 videos for training, 16 sequences with 120 frames each for validation and 16 sequences with 120 frames each for testing. Note that the ground-truth of testing data are not released. Therefore, in this paper, we only show the validation results for Vid3oC dataset. In order to evaluate the validity of our network, we choose 10 videos (050 to 059) from another dataset, IntVID \cite{vid3oc}, as a secondary test dataset. For each video, we extract 14 consecutive frames for testing.  

\subsection{Implementation Details}

In the proposed DKSAN networks, to compare with EDVR, we set the kernel size as $3 \times 3$ with 128 filters for most of Conv layers, all deformable kernel layers and all deformable convolution layers. The kernel size of feature fusion layers is $1 \times 1$. The reduction ratio of channel attention module is still $r = 16$ as \cite{RCAN}. 5 resblocks are in feature extractor. The number of RCAB blocks are set to 30, 20, 15, 10 for each level (from 1 to 4) of reconstruction module. The PixelShuffle layer is the same as \cite{ESPCNN}. The last Conv layer filter is set to 3 in order to output color frames.

\begin{figure}
    \centering
    \includegraphics[width=11cm]{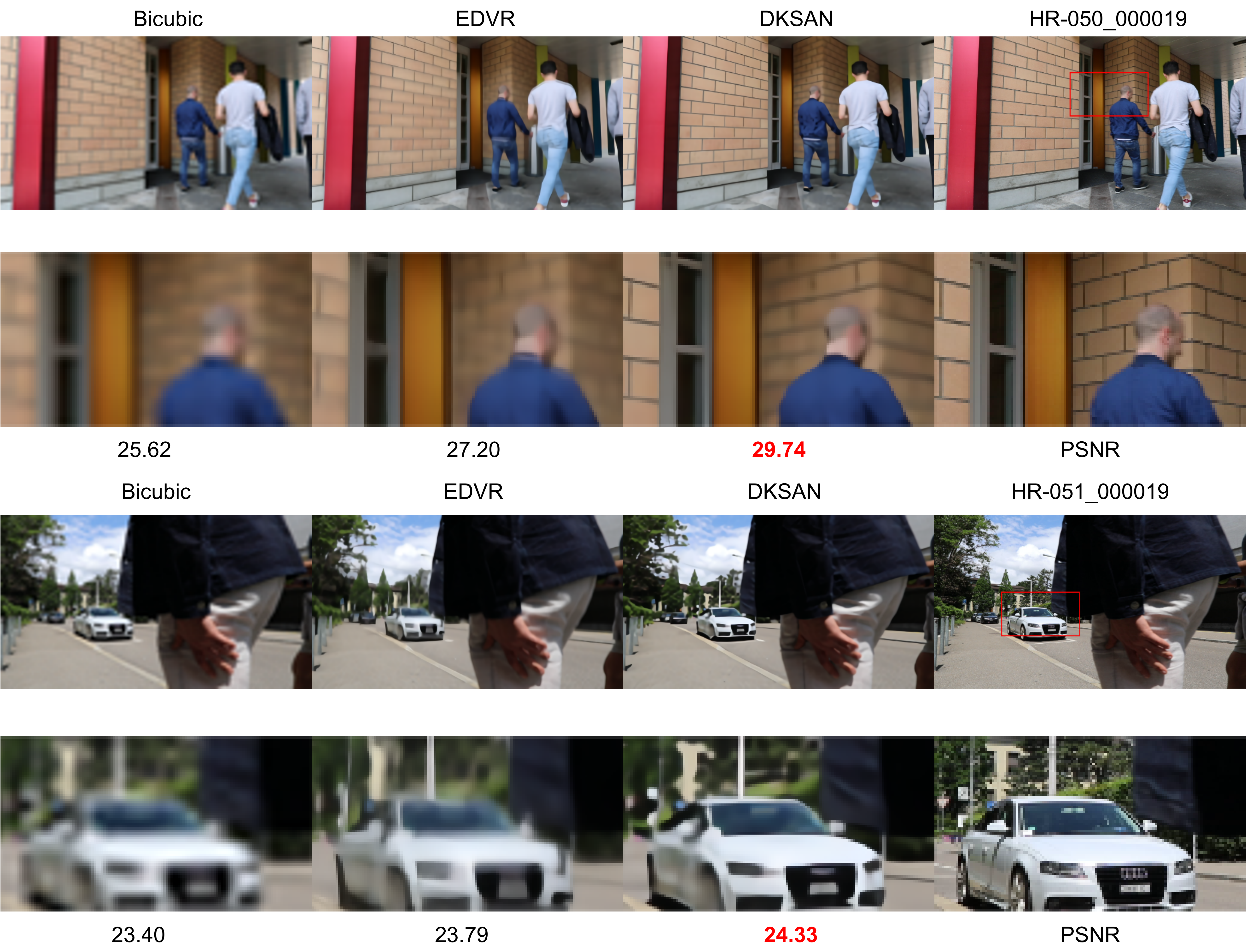}
    \caption{Visual comparison results among competing approaches for Vid3oC dataset (video 050 and 051) at a scaling factor of 16.}
    \label{fig:ch4_psnr3}
\end{figure}

In particular, we randomly crop the 7 low-res frames as small patches with the size of $32 \times 32$, and crop the corresponding 4th high-res frames with the size of $512 \times 512$. The batch size is 16. We augment the training set by random flips and rotations. The optimizer we used is ADAM \cite{adam} with  $ \beta_1 = 0.9 $, $ \beta_2 = 0.999 $. The initial learning rate is set to \num{4e-4}. The total training step is $115k$. The loss function we used is adapted Charbonnier penalty function \cite{laplace}. The loss can be defined as eqn.~\ref{eqn:char_loss} shown as follows:

\begin{equation}
    Loss = \sqrt{||\hat{Y}_r - Y_r||^2+\xi^2}
\label{eqn:char_loss}
\end{equation}
where $\xi =$ \num{1e-3}, $\hat{Y}_r$ is super-resolved frame and $Y_r$ is target frame (ground-truth). All experiments are trained on 4 NVIDIA Titan Xp GPUs with PyTorch framework Implementation.

Note that for fair comparison, we retrain EDVR with the same training dataset (Vid3oC) and keep most of EDVR setting as the same as the original implementation to run the experiment except setting the upscale module from factor 4 to factor 16 in order to make sure EDVR can generate extreme super-resolved frames.

\subsection{Comparison Against State-of-the-Art}

Because few existing works related to video extreme super-resolution (with a scale factor of 16), in this work, we have compared our proposed network against with Bicubic interpolation and state-of-the-art EDVR. 

Table~\ref{tab:PSNR_vid3oc} shows the PSNR comparison results, number of parameters and running time (seconds per frame) of our approach with the competing methods, Bicubic interpolation and EDVR with the scaling factor of 16 on the validation set of Vid3oC \cite{vid3oc}. From the Table, we can see that our DKSAN method has the best PSNR scores for all 16 testing videos. The significant PSNR gains (up to $2.4dB$) over previous state-of-the-art method EDVR. Since PSNR metrics cannot always evaluate the subjective quality of images, therefore, a qualitative result is shown in Fig.~\ref{fig:ch4_psnr3}, we can easily observe that our proposed network DKSAN can better reconstruct the lines on the wall for ``050\_000019'' and a clearer car for ``051\_000019'' compared with EDVR. 

\begin{figure}
    \centering
    \includegraphics[width=11cm]{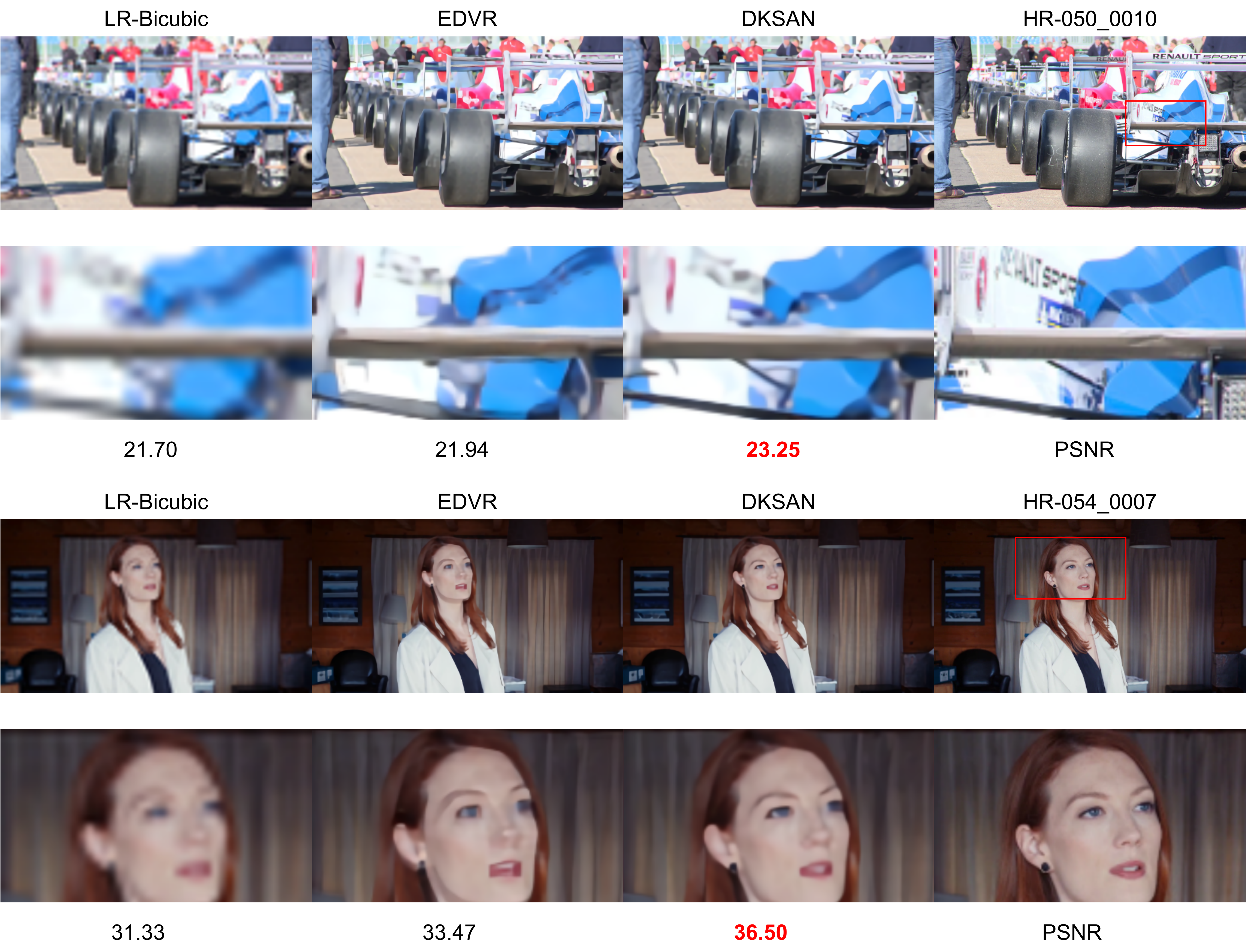}
    \caption{Visual comparison results among competing approaches for IntVID dataset (video 050 and 054) at a scaling factor of 16.}
    \label{fig:ch4_psnr2}
\end{figure}

To further verify the effectiveness of our proposed method, we selected another dataset, IntVid \cite{vid3oc} as a secondary test dataset. From Table~\ref{tab:PSNR_ch4}, we can easily find out that our proposed DKSAN has the best performance for all 10 testing videos compared with EDVR and bicubic interpolation. A qualitative result is shown in Fig.~\ref{fig:ch4_psnr2}. For the subject ```050\_0010'', compared with EDVR, our DKSAN can better recover more details of the rear wing. Taking another example, in ``054\_0007'', our DKSAN can reconstruct a much clearer face than EDVR does.

\subsection{Ablation Studies}

To investigate the effect of proposed DKC\_Align module and DKSA module, we have conducted different strategies to remove the certain components from the final framework DKSAN. In particular, we have implemented four competing models for our ablation studies: 1) training with only resblocks, without channel attention, alignment and DKSA; 2) training without DKC\_Align and DKSA module; 3) training with DKC\_Align module but without DKSA module; 4) training with all modules (proposed DKSAN). Note that all experiments are trained under same dataset and conditions for fair comparison.

Table~\ref{tab:ab_ch4} shows the results, the number of parameters and running time (seconds per frame) of all four strategies mentioned previously with PSNR scores of each video and average. The backbone result is running only based on resblock, no channel attention, alignment and DKSA applied. From the results, we can see that the backbone has the worst performance; adding channel attention module but without DKC\_Align and DKSA modules, the result is only 31.27 dB; after adding DKC\_Align module, the result is improved to 31.32 dB; finally, we observe that after adding DKSA module (the full version of DKSAN), the result is further improved to 31.43 dB (\textbf{0.4dB} and \textbf{0.16dB} gained when compared with backbone and w/o Alignment \& DKSA respectively) because of the effective module DKSA. 

\begin{table}
    \begin{center}
    \begin{tabular}{c|c|c|c|c}
    \hline
    \multirow{2}{*}{Video Name}  & Backbone & w/o Alignment \& DKSA  & w/o DKSA & DKSAN (ours)  \\
    \cline{2-5} & PSNR (dB) & PSNR (dB) & PSNR (dB) & PSNR (dB) \\
    \hline\hline
      050  & 22.80 & 22.98   & 22.87   & \textbf{23.06}   \\
      051  & 24.72 & 24.89   & 24.89   & \textbf{24.92}  \\
      052  & 31.65 & 31.75   & 31.85   & \textbf{31.87}  \\
      053  & 25.05 & 25.07   & \textbf{25.18}   & 25.09 \\
      054  & 35.30 & 35.73   & 35.86   & \textbf{36.18}  \\
      055  & 26.52 & \textbf{26.89}   & 26.69   & 26.88  \\
      056  & 33.79 & 33.92   & \textbf{34.33}   & 34.22  \\
      057  & 38.97 & 39.13   & 39.27   & \textbf{39.75} \\
      058  & 38.09 & 38.21   & \textbf{38.30}   & 38.15 \\
      059  & 33.38 & 34.15   & 34.16   & \textbf{34.17} \\
      \hline\hline
    Average & 31.03 & 31.27   & 31.32   &  \textbf{31.43} \\
    \hline
    Parameters & 26.1M  &  26.3M  & 27.0M  & 29.5M \\
    \hline
    Runtime(s/f) & 0.83 & 0.89  &  0.92  & 0.97 \\
 
    \hline
    \end{tabular}
    \end{center}
    \caption{Ablation Studies for DKSAN on IntVID dataset for scaling factor of 16. Backbone means only resblocks used; w/o Alignment \& DKSA means DKC\_Align and DKSA Module are not applied; w/o DKSA means only the DKSA module is not applied; s/f means seconds per frame. \textbf{Bold} font indicates the best result.}
    \label{tab:ab_ch4}
\end{table}

\subsection{AIM 2020 Video Challenge}

We have participated in the AIM2020 video extreme super-resolution challenges which is the second edition of AIM2019 challenges \cite{AIM2019}. Our submissions won the \textbf{2nd place} for both track 1 and track 2 competitions. Note that track 1 is based on PSNR performance and track 2 is based on perceptual (see the AIM2020 challenge report \cite{AIM2020} for more details).

\section{Conclusions}

In this work, we proposed a multi-frame based VSR networks DKSAN for extreme low-resolution videos. The novel temporal alignment module, DKC\_Align, can help the networks to better learn and align the detailed features by improving both theoretical and effective receptive fields between reference frame and its neighboring frames. Furthermore, the DKSA module calibrated the reconstructed features to further enhance the edges and textures at the spatial domain. Thanks to the newly designed DKC\_Align and DKSA modules, the proposed architecture can reconstruct high-quality HR frames from extreme LR frames and significantly improve both objective and subjective performance when compared with state-of-the-art approach EDVR \cite{EDVR}. 

\section*{Acknowledgment}

This work is partially supported by the NSF under grants IIS-1908215 and OAC-1839909, the DoJ/NIJ under grant NIJ 2018-75-CX-0032, and the WV
Higher Education Policy Commission Grant (HEPC.dsr.18.5).



%
%

\end{document}